\documentclass{article}

\PassOptionsToPackage{numbers, compress}{natbib}

\usepackage[preprint]{neurips_2026}

\usepackage[utf8]{inputenc} 
\usepackage[T1]{fontenc}    
\usepackage{hyperref}       
\usepackage{url}            
\usepackage{booktabs}       
\usepackage{amsfonts}       
\usepackage{nicefrac}       
\usepackage{microtype}      
\usepackage{xcolor}         

\usepackage[utf8]{inputenc} 
\usepackage[T1]{fontenc}    
\usepackage{hyperref}       
\usepackage{url}            
\usepackage{booktabs}       
\usepackage{amsfonts}       
\usepackage{nicefrac}       
\usepackage{microtype}      
\usepackage{xcolor}         

\usepackage{xcolor}   
\usepackage[dvipsnames]{xcolor}
\usepackage{booktabs}
\usepackage{amsmath}
\usepackage{makecell}
\usepackage{wrapfig}
\usepackage{enumitem}  

\usepackage{multirow}
\usepackage{graphicx} 
\usepackage[table]{xcolor}
\usepackage{tcolorbox}
\usepackage{booktabs}
\usepackage{pifont}

\title{
Reading Cognition as Decisions Unfold in Words: \\
A Factorized Inverse Decision Model
}

\author{%
\textbf{Jiawen Kang}\textsuperscript{1}
\qquad
\textbf{Dongrui Han}\textsuperscript{1}
\qquad
\textbf{Xixin Wu}\textsuperscript{1}
\qquad
\textbf{Helen Meng}\textsuperscript{1}
\\[0.4em]
\textsuperscript{1}The Chinese University of Hong Kong
\\
\texttt{\{jwkang\}@se.cuhk.edu.hk}
}

\begin{document}

\maketitle

\begin{abstract}

Inverse decision modeling infers latent properties of decision processes from observed behavior, but existing formulations rely primarily on action trajectories. 
In verbalized cognitive tasks, task execution also produces response dynamics that action-only formulations leave unmodeled, such as verbal production, interaction, and hesitation.
We propose a \textit{factorized inverse decision model (FIDM)} that decomposes each individual's task-execution likelihood into an \textit{action factor} and an \textit{effort factor}, governed by separate individual-specific parameters.
From raw verbal transcripts, a language model produces structured task-execution traces for factorized inference.
On data from 400 older adults performing a grocery-shopping dialog task for cognitive screening, controlled recovery shows selective estimation of the intended factors, while matched semi-synthetic conditions show that FIDM preserves action-execution distinctions even when aggregate behavioral summaries are matched.
Action evidence further localizes task-defined deviations across participants.
In cognitive-status classification, FIDM provides information complementary to clinical scores, trajectory summaries, and frozen language representations, with consistent gains across all evaluated baselines in the binary setting.
\end{abstract}
\section{Introduction}
\vspace{-5pt}

Human decision behavior exhibits rich variability:
when driving home, one may cut quickly through busy intersections or cruise along a quiet waterfront; a football attack may drive straight at goal or unfold through patient passes and detours.
Such variations reflect the underlying cognitive processes of decision-makers, including interpreting goals, forming plans, selecting actions, and organizing their execution.
Inferring properties of the underlying decision process from observed behavior is a central problem in fields like cognitive science and behavioral economics \cite{gergely1995taking,dellavigna2009psychology,krajbich2010visual}.
In this work, we model decision trajectories from a verbalized cognitive task to infer interpretable participant descriptors and characterize cognitive status.

\begin{figure}[tbp]
\begin{center}
\includegraphics[width=0.7\linewidth,scale=1.0]{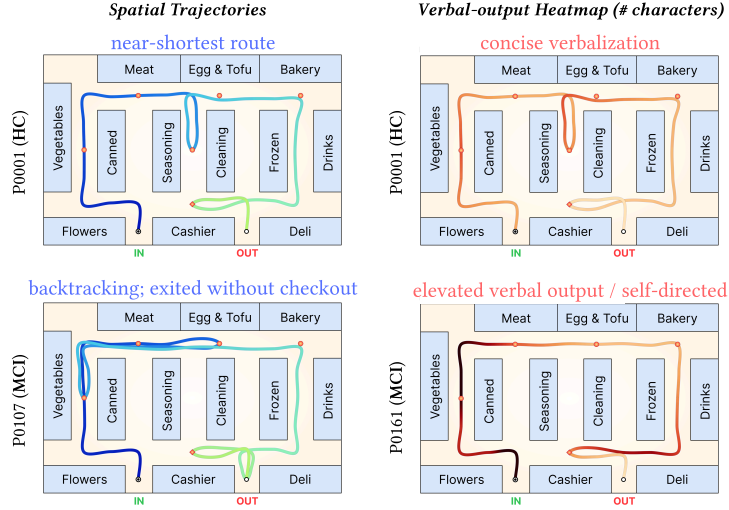}
\end{center}
\vspace{-10pt}
\caption{
\small
Left: spatial
trajectories show that participant P0001 (Health Control, HC) follows a
near-shortest route, while participant P0107 (Mild Cognitive Impairment, MCI) backtracks
and exits without checkout. Right: transition-aligned verbal output, derived from character counts of route descriptions, shows that
P0001 verbalizes concisely, while participant P0161 (MCI)
produces elevated verbal effort with self-directed
verbalization.
}
\vspace{-15pt}
\label{fig:fig1}
\end{figure}

Inverse decision modeling (IDM) offers a unifying, model-based perspective for interpreting observed decision behavior \cite{idm0,idm1}.
IDM specifies how a decision process generates behavior and reasons backward from observations to the factors that shaped it.
The forward process is typically formulated as a Markov decision process that structures states, actions, and transitions under reinforcement learning or planning models.
Different instantiations recover different latent quantities: inverse reinforcement learning recovers reward functions from demonstrated behavior \cite{irl01,dirl}, Bayesian inverse planning infers goals or beliefs underlying observed actions \cite{ip,chandra2023inferring}, and inverse constraint learning identifies constraints that shape feasible behavior \cite{icrl}.

Existing IDM formulations base inference primarily on action trajectories.
In cognitive assessment, however, task execution performance reflects both the actions participants select and the response dynamics that accompany each choice.
This distinction is especially pronounced when a task requires coordinating multiple steps toward a goal: two participants may select similar actions yet differ markedly in their fluency, hesitation, and need for external guidance at each step \cite{shallice1991deficits,gsdt}.
Action-only formulations therefore capture only part of task performance, leaving response dynamics unmodeled.

To address this, we propose a factorized inverse decision model for verbalized cognitive task execution. 
The model factorizes the likelihood of each participant's task execution into an action factor and an effort factor, which capture variation in action choices and in response dynamics, respectively. Because separate participant-specific parameters govern each factor, the factorization preserves individual differences that action-only formulations conflate—two participants who choose identically can still yield distinct effort profiles.
From raw verbal transcripts, a large language model produces the structured task-execution trajectories, which the factorized model then analyzes.

We collect verbal task-execution data from 402 older adults performing
a grocery-shopping dialog task designed for cognitive screening~\cite{gsdt},
in which participants describe how they would navigate a store to collect
ingredients for a dish. Our contributions are as follows:
\begin{itemize}[leftmargin=*, topsep=2pt, itemsep=2pt]
    \item We introduce FIDM, which separates how an individual selects actions from how those actions are executed, yielding an interpretable action--effort profile from verbalized task-execution traces.
    \item We validate selective recovery of the action–effort decomposition and show that FIDM distinguishes matched task-execution conditions that aggregate behavioral summaries do not.
    \item Using cognitive-screening data collected from 402 older adults, we show that FIDM provides interpretable, task-localized evidence and complementary information beyond conventional clinical and data-driven representations.
\end{itemize}


\vspace{-10pt}
\section{Related Work}
\vspace{-5pt}

\subsection{Inverse Decision Modeling}
\vspace{-5pt}

Inverse decision modeling (IDM) provides a common formulation for inferring latent properties of a decision process from observed behavior. 
Related approaches differ mainly in the quantities they infer and the behavioral evidence they use. Our work fixes the task model and infers individual-specific action and effort parameters from action choices and response dynamics~\cite{idm1}.

\noindent \textbf{Inverse reinforcement learning.}
Inverse reinforcement learning recovers a reward function under which observed state–action trajectories are optimal or near-optimal~\cite{irl0,arora2021survey}.
Maximum-entropy IRL casts this inference as a probabilistic model over complete trajectories ~\cite{irl01}.
Subsequent work relaxes individual components of this formulation: inverse constrained reinforcement learning treats task constraints as unknown while retaining a nominal reward~\cite{icrl}, dynamic IRL allows rewards to vary over time~\cite{dirl}, and robust Bayesian formulations accommodate uncertainty in the agent’s model of environment dynamics~\cite{wei2023bayesian}.
These approaches primarily infer properties of the task or objective; we instead hold these fixed and infer descriptive differences across individuals.

\noindent \textbf{Bayesian inverse planning.}
Bayesian inverse planning infers latent mental states by inverting a model of goal-directed action.
Early models estimate goals and beliefs by comparing observed actions with those expected from approximately rational agents~\cite{ip,baker2017rational}.
Later work explicitly models planning limitations, including bounded search, partial plans, replanning, and failed actions~\cite{zhi2020online}.
Recent systems integrate multimodal observations and language-model-driven policy approximation into Bayesian inverse planning for complex household environments~\cite{jin2024mmtom,zhang2025overcoming}.
Our work differs in holding the task goal fixed and instead inferring individual-level action and effort characteristics.

\noindent \textbf{Evidence accumulation models.}
Evidence-accumulation models leverage response time alongside choice to characterize the decision process.
The drift-diffusion model accounts for both through latent parameters governing evidence accumulation, decision boundaries, and non-decision time~\cite{ratcliff2008diffusion}.
Reinforcement-learning diffusion models further connect learned values to the accumulation process~\cite{pedersen2017drift,fontanesi2019reinforcement}; incorporating response times has also been shown to improve the reliability of decision-model parameter estimates~\cite{shahar2019improving}.
These models typically describe trial-level decisions among a small number of alternatives, whereas our setting involves response dynamics across an extended sequence of interdependent verbalized decisions.

\subsection{Cognitive Assessment}

Cognitive assessment uses structured tasks to probe specific cognitive abilities through observable behavior. We review two directions relevant to our work: (i) speech-based assessment, and (ii) planning and navigation assessment.

\noindent \textbf{Speech-based cognitive assessment.}
Speech-based assessments elicit language under varying degrees of task structure to examine cognitive and communicative abilities.
Standardized picture-description tasks such as Cookie Theft probe object naming and scene-level information conveyance~\cite{giles1996performance}.
Narrative and story-retelling tasks further engage event sequencing and discourse organization, while interviews and conversations elicit less constrained speech, revealing fluency and interaction patterns~\cite{boschi2017connected}.

\noindent \textbf{Planning and navigation assessment.}
Planning and navigation assessments examine how individuals coordinate multistep, goal-directed behavior in structured everyday tasks.
The Multiple Errands Test evaluates the coordination of several goals and rules in a real-world setting~\cite{shallice1991deficits}, while the Naturalistic Action Test assesses the completion and errors of multistep daily activities~\cite{schwartz2002naturalistic}.
Virtual multiple-errands and supermarket tasks impose comparable demands within controlled environments to assess executive function, memory, and spatial navigation~\cite{rand2009validation,yan2021virtual}.
The Hong Kong Grocery Shopping Dialog Task embeds grocery-shopping planning and route navigation in spoken interaction, and serves as the task setting for the present work~\cite{gsdt}.


\vspace{-5pt}
\section{Factorized Inverse Decision Model}
\label{sec:method}
\vspace{-5pt}

We formalize verbalized cognitive task execution as an observed sequence under a shared task model and describe a factorized inverse decision model that infers individual-specific descriptors from this sequence.

\subsection{Problem Formulation}
\label{sec:problem-formulation}
\vspace{-5pt}
Let \(M = (S, A, T, r, \gamma)\) be a Markov decision process representing the structured sequential task, 
where \(S\) is the state space, \(A\) is the action space, \(T(s' \mid s,a)\) is the transition kernel, \(r(s,a)\) is the task reward, and \(\gamma\) is the discount factor. 
The task model \(M\) remains fixed across individuals and provides the common sequential-decision structure for the inverse model.

A conventional decision trajectory records selected actions and resulting state transitions. 
In a verbalized cognitive task, each step additionally carries signals reflecting how the action was executed. 
We therefore represent individual \(j\)'s task execution as a trace
\begin{equation}
\label{eq:task-trace}
    \zeta_j
    =
    \bigl(
        (s_i, a_i, e_i, s_{i+1})
    \bigr)_{i=0}^{L_j-1},
\end{equation}
where \(s_i \in S\) is the current task state, \(a_i \in A\) is the selected action, \(e_i\) is the observable execution signal aligned with the step, and \(s_{i+1} \in S\) is the resulting state. The execution signal may be scalar or vector-valued, encompassing measurements such as response latency, verbal production, and revision behavior. Throughout, we condition on the observed initial state \(s_0\) and trace
length \(L_j\), which are omitted from the likelihood notation.
For notational simplicity, action and execution observations share an index here; task-specific instantiations may align them at different granularities.

We characterize individual \(j\)'s task execution through the descriptive parameters
\begin{equation}
    \phi_j = (\eta_j, \theta_j),
\end{equation}
where \(\eta_j\) governs the individual's action choices and \(\theta_j\) governs the conditional distribution of the execution signals. Given the shared task model and an observed task-execution trace, the inverse problem is
\begin{equation}
    (M, \zeta_j)
    \longrightarrow
    \phi_j.
\end{equation}

Thus, \(M\) specifies the task structure shared across individuals, while \(\phi_j\) captures individual-specific variation in task execution. The following sections define how \(\eta_j\) and \(\theta_j\) enter the action and effort factors of the trajectory likelihood.

\begin{figure}[tbp]
\begin{center}
\hspace{-30pt}
\includegraphics[width=1.1\linewidth,scale=1.0]{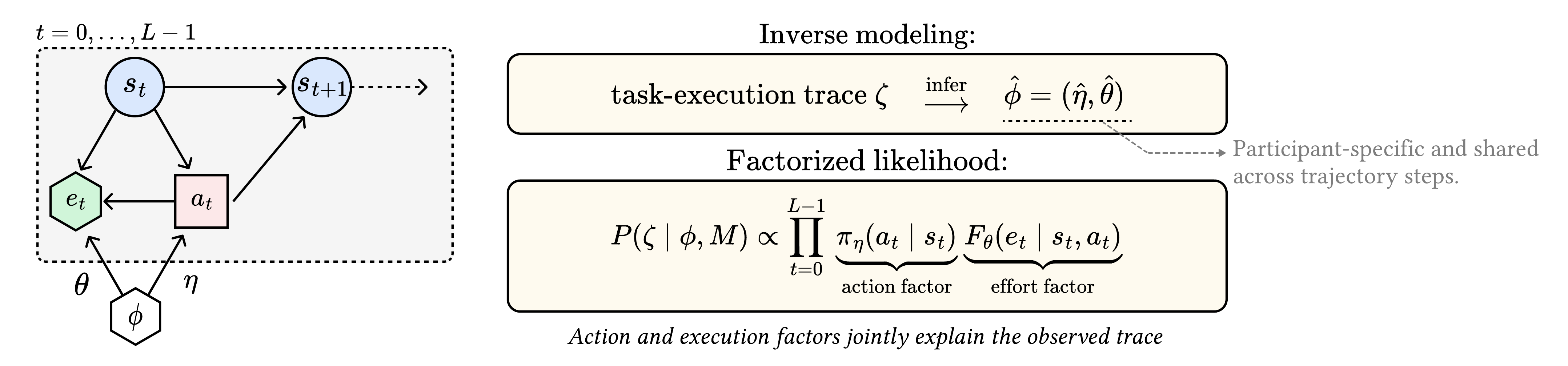}
\end{center}
\vspace{-10pt}
\caption{
\small
Graphical model (left) and factorized likelihood (right) of the proposed inverse decision model.
The action factor models choices under participant-specific parameter $\eta$, while the effort factor models execution signals conditional on the corresponding task execution under participant-specific parameters $\theta$.
Both factors are defined over the shared task model $M$.
}
\vspace{-10pt}
\label{fig:graph}
\end{figure}

\subsection{Factorized Task-Execution Model}
\label{sec:factorized-model}
\vspace{-5pt}
Given the shared task model $M$, we define a forward model for the action choices and execution signals observed along an individual's task-execution trace.

\noindent\textbf{Shared task reference.}
We first derive a shared action-value function from the task
specification using the soft Bellman equations
\begin{equation}
\begin{aligned}
Q(s,a)
&=
r(s,a)
+
\gamma\sum_{s'}T(s'\mid s,a)V(s'),
\\
V(s)
&=
\tau_Q
\log\sum_{a'\in A(s)}
\exp\!\left(\frac{Q(s,a')}{\tau_Q}\right).
\end{aligned}
\end{equation}

The reward function, discount factor, and Bellman temperature are
fixed globally and shared across individuals. The resulting $Q(s,a)$
therefore provides a common task-value reference for evaluating
individual differences in action selection.

\noindent\textbf{Action factor.}
Individual $j$'s action probabilities are defined by
\begin{equation}
\pi_{\eta_j}(a\mid s)
=
\frac{
\exp\!\left(\eta_jQ(s,a)\right)
}{
\sum_{a'\in A(s)}
\exp\!\left(\eta_jQ(s,a')\right)
},
\qquad
\eta_j\geq 0.
\end{equation}

The parameter $\eta_j$ controls the sensitivity of action probabilities to task-value differences. Lower values distribute probability more evenly across the available actions, while larger values concentrate probability on actions with higher $Q$-values. 

\noindent\textbf{Effort factor.}
For each realized state--action step, the aligned execution signal is modeled as
$e_i\sim F_{\theta_j}(\cdot\mid s_i,a_i)$.
The parameter $\theta_j$ captures individual-specific variation in this conditional distribution. The concrete form of $F_{\theta_j}$ and the execution channels included in $e_i$ are specified in the task instantiation.

\noindent\textbf{Factorized likelihood.}
Combining the action factor, effort factor, and shared task transition gives the trace likelihood:
\begin{equation}
\label{eq:factorized-likelihood}
P(\zeta_j\mid\phi_j,M)
=
\prod_{i=0}^{L_j-1}
\underbrace{
\pi_{\eta_j}(a_i\mid s_i)
}_{\text{action factor}}
\underbrace{
F_{\theta_j}(e_i\mid s_i,a_i)
}_{\text{effort factor}}
T(s_{i+1}\mid s_i,a_i).
\end{equation}

The transition kernel is fixed by the shared task model. Removing terms that are constant with respect to the individual-specific parameters yields
\begin{equation}
\mathcal{L}_j(\eta_j,\theta_j)
=
\sum_{i=0}^{L_j-1}
\log\pi_{\eta_j}(a_i\mid s_i)
+
\sum_{i=0}^{L_j-1}
\log F_{\theta_j}(e_i\mid s_i,a_i).
\end{equation}

The two terms connect complementary observations to the two individual-specific parameter blocks: the realized state--action sequence provides evidence for \(\eta_j\), and the aligned execution signals provide evidence for \(\theta_j\) conditional on that sequence. 
This separation allows individuals with similar action paths to differ in their inferred execution-effort profiles.

\noindent\textbf{Individual-level estimation.}
Given the observed trace \(\zeta_j\), we estimate the individual-specific parameters by maximum likelihood:
\begin{equation}
\hat{\phi}_j
=
\arg\max_{\eta_j\geq 0,\,\theta_j}
\mathcal{L}_j(\eta_j,\theta_j),
\qquad
\hat{\phi}_j=(\hat{\eta}_j,\hat{\theta}_j).
\end{equation}
Under the factorized objective, the action and effort terms separately determine \(\hat{\eta}_j\) and \(\hat{\theta}_j\), respectively. The resulting \(\hat{\phi}_j\) forms an individual-specific, model-based profile of task execution. 
Clinical outcomes and other external measurements are used only in subsequent analyses and do not enter the estimation objective.





\subsection{Interpretation}
\label{sec:factor-interpretation}
\vspace{-5pt}

The two factors admit a common context-calibrated interpretation under the
shared task model. 
For any state \(s\) and two available actions \(a,a'\in A(s)\), the fitted policy satisfies
\begin{equation}
\log
\frac{\pi_{\eta}(a\mid s)}
     {\pi_{\eta}(a'\mid s)}
=
\eta\bigl[Q(s,a)-Q(s,a')\bigr].
\end{equation}
Thus, \(\hat{\eta}_j\) is determined by how the observed choices trade off
relative task values among the alternatives available at the same visited
states. It differs from an error count or route-length summary because the
evidence associated with a choice depends on its \(Q\)-gap to the available
alternatives.

The effort parameters apply the same conditional principle at the execution
level. An observed signal is interpreted relative to the effort distribution
expected under its aligned execution context, rather than by its raw magnitude
alone. Identical duration or verbosity can therefore carry different evidence
under different task demands. In this sense, the inferred factors separate
value-relative action selection from context-relative execution effort.
\section{Model Instantiation for the Grocery-Shopping Task}
\label{sec:task-instantiation}
\vspace{-5pt}

We instantiate FIDM on the navigation-and-purchase component of the Hong Kong Grocery Shopping Dialog Task (HK-GSDT)~\cite{gsdt}.
The shared task model specifies the supermarket environment and task objectives, while timestamped dialog is grounded into the action and execution observations used for individual-level inference.

\subsection{Task Structure}
\label{sec:grocery-task-structure}

The HK-GSDT requires participants to verbally direct navigation through a fixed supermarket layout while collecting target items.
We represent the layout as a discrete grid world, with each state recording the current location, task progress, and active task phase.
The task proceeds through initial shopping, an assessor-introduced bakery request, and checkout.
Each phase activates the corresponding task objective, yielding phase-dependent action values under the shared task model.
Because the environment and task objectives are fixed across participants, this construction provides the common task-value reference used by the action factor.
Further details of the task representation are provided in Appendix~\ref{app:task-details}.

\subsection{Transcript-to-Trace Construction}
\label{sec:trace-construction}
We convert each timestamped participant--assessor dialog into an ordered movement trace and aligned execution observations.
A language model grounds each route-description segment to a sequence of grid movements using the preceding dialog context and current location; deterministic replay then reconstructs the visited states, while grounded purchase and payment events update task progress.

In this instantiation, grid transitions and speech segments are not in one-to-one correspondence: a route-description segment may encode several consecutive movements.
The action likelihood is evaluated over grounded movements and task-completion actions, whereas execution observations are measured at the route-description-segment level and conditioned on the number of aligned movements.
For each segment, the four execution channels are participant speech duration, character count, dialog-turn count, and positive pause duration.
We use Gamma, negative-binomial, negative-binomial, and Bernoulli–log-normal likelihoods, respectively; exact parameterizations and grounding details are given in Appendix~\ref{app:task-details}.

\vspace{-5pt}
\section{Experiments}
\vspace{-5pt}

\subsection{Experimental Settings}
\label{sec:experimental-settings}
\vspace{-5pt}

\noindent\textbf{Data.}
We use 400 participants with valid HK-GSDT task-execution traces after excluding two participants who did not complete the assessment.
The predefined participant-level split contains 332 training and 68 held-out test participants.

\noindent\textbf{Representations.}
FIDM yields five parameters per participant: one action-sensitivity parameter and four effort parameters corresponding to speech duration, character count, dialog turns, and positive pause duration.
Population-level likelihood parameters are estimated on the training partition, after which individual-specific parameters are inferred from each participant's trace.
We compare FIDM with the HK-GSDT total score, the total score plus seven subscores, 14 trajectory-level summary statistics, and frozen Chinese BERT and MacBERT representations computed from participant-only utterances.
Full feature definitions, encoder configurations, model links, pooling, and fusion procedures are provided in Appendix~\ref{app:experimental-details}.

 
 
\noindent\textbf{Classification setup.}
We use the same class-balanced logistic-regression classifier for all representations so that performance differences primarily reflect the representation.
The primary task is binary NCD classification, grouping MCI and major NCD against normal cognition; we additionally report three-class classification among normal cognition, MCI (Mild Cognitive Impairment), and major NCD.
We report five-fold stratified cross-validation on the training partition and a single evaluation on the held-out test partition.
Binary results use ROC-AUC and balanced accuracy; three-class results use macro one-vs-rest ROC-AUC and balanced accuracy.
All preprocessing is fitted without access to the held-out test data; implementation details are given in Appendix~\ref{app:experimental-details}.



\subsection{Validating the Factorization}
\label{sec:factor-recovery}

We evaluate whether the two factors separately recover variation in action choice and execution behavior.

\begin{figure}[tbp]
\begin{center}
\hspace{-0pt}
\includegraphics[width=1.0\linewidth,scale=1.0]{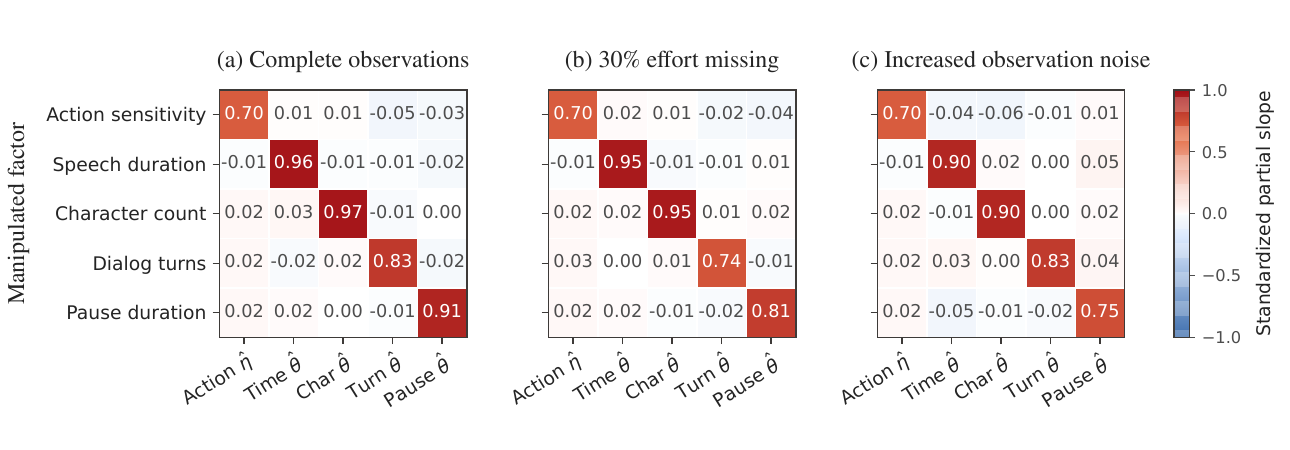}
\end{center}
\vspace{-20pt}
\caption{
\small
Controlled factor recovery on semi-synthetic HK-GSDT traces.
Rows denote independently manipulated generative factors, columns denote recovered parameters, and each cell reports the standardized partial slope controlling for the remaining manipulations. 
The diagonal structure is preserved under (a) complete observations, (b) \(30\%\) missing effort observations, and (c) increased observation noise.
}
\vspace{-10pt}
\label{fig:heatmap}
\end{figure}

\noindent\textbf{Controlled recovery.}
We first test whether the inferred parameters recover the behavioral dimensions they target. 
Using the observed HK-GSDT task structure, we generate semi-synthetic traces by independently varying action sensitivity, speech duration, character count, dialog-turn count, and positive pause duration.
Action sequences are sampled from the task-value policy, while effort observations are generated at the segment level; all parameters are then re-estimated using the same inference procedure as for the observed data.
Figure~\ref{fig:heatmap} shows a strongly diagonal response matrix under complete observations, with matched responses ranging from \(0.70\) to \(0.97\) and cross-factor responses remaining near zero.
This structure is preserved under 30\% missing effort observations and increased observation noise, indicating selective recovery of action-side and effort-side variation with limited cross-loading.


\noindent\textbf{Matched observable behavior.}
We next test whether conditioning execution on the realized action sequence captures distinctions beyond aggregate behavioral summaries. For example, a
participant may take a longer route while executing each segment efficiently, whereas another may take a shorter route but require more time, verbal output, or interaction per segment; their aggregate
\begin{wrapfigure}{r}{0.35\textwidth}
\vspace{-15pt}
\label{fig:auc}
\vspace{10pt}
\includegraphics[width=\linewidth]{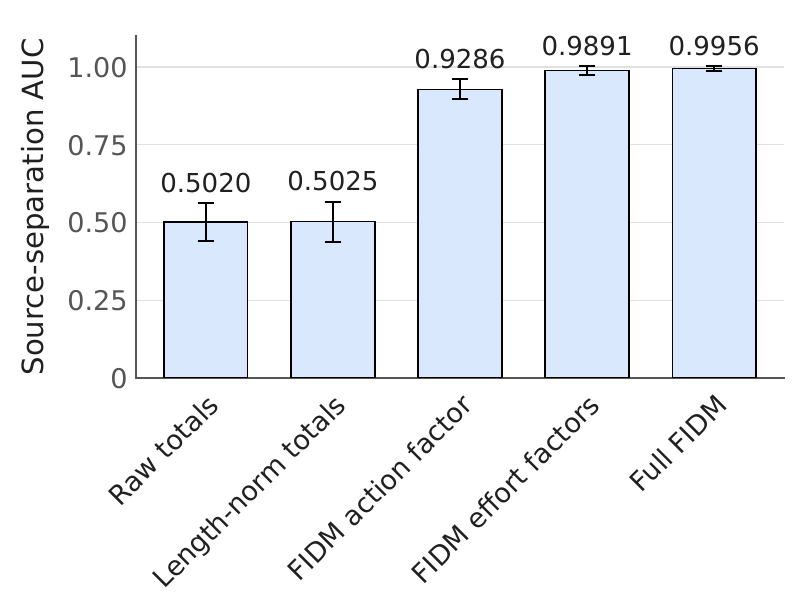}
\vspace{-20pt}
\caption{
AUC for distinguishing the two matched semi-synthetic conditions using aggregate summaries and FIDM features.
}
\vspace{-15pt}
\label{fig:auc}
\end{wrapfigure}
\hspace{-5pt}
execution statistics can nevertheless be similar.
We therefore construct paired semi-synthetic traces from the same observed HK-GSDT state--value contexts and with identical movement exposure, while assigning the two members different action and effort parameters.
Raw and per-movement execution summaries are matched in expectation across the two groups (maximum absolute standardized mean difference \(=0.072\)).

Figure~\ref{fig:auc} compares the resulting representations.
The raw and per-movement summaries remain at chance (AUC \(=0.502\) and \(0.503\)), whereas the FIDM action factor, effort factors, and full representation achieve AUCs of \(0.929\), \(0.989\), and \(0.996\), respectively.
Thus FIDM distinguishes the two task-execution conditions even when their aggregate behavioral summaries are matched.

\subsection{Localizing Action and Effort Evidence}
\label{sec:local-factor-evidence}
\vspace{-5pt}

We next examine where along individual task executions the two likelihood factors receive their evidence.
For action, we define unexpectedness as the negative local log-likelihood of each grounded movement under the inferred policy. For effort, we compute residuals that measure segment-level effort relative to its conditional expectation given the number of grounded movements.

\begin{figure}[h]
\begin{center}
\includegraphics[width=1\linewidth,scale=1.0]{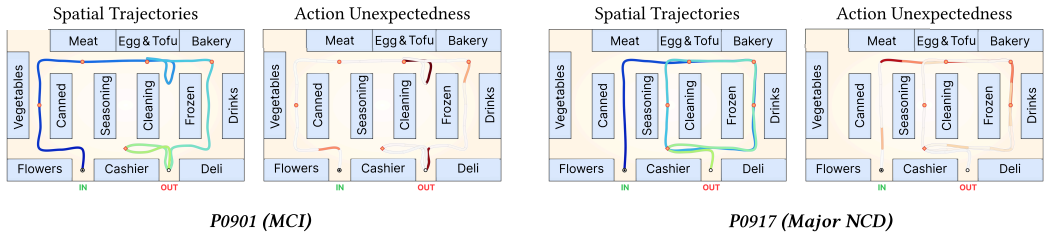}
\end{center}
\vspace{-10pt}
\caption{
\small
Local action evidence for P0901 (MCI) and P0917 (Major NCD). For each participant, the grounded trajectory is shown alongside step-level action unexpectedness under the fitted participant-specific policy; darker highlighting indicates greater unexpectedness.
}
\label{fig:sample-action}
\end{figure}

\noindent\textbf{Action evidence.}
Figure~\ref{fig:sample-action} shows two examples of localized action evidence.
For participant P0901, actions with elevated unexpectedness coincide with erroneous turns and the subsequent backtracking needed to recover the route.
For participant P0917, elevated action unexpectedness occurs when the participant passes the meat section without completing the expected purchase.
The latter is not an overt navigation error, yet it is locally unexpected under the current task objective.
Together, these examples illustrate that action unexpectedness can localize atypical task choices ranging from geometric route deviations to missed task opportunities.

\begin{figure}[h]
\begin{center}
\includegraphics[width=1\linewidth,scale=1.0]{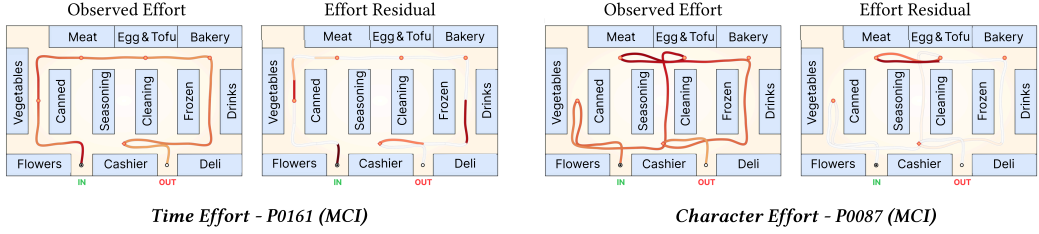}
\end{center}
\vspace{-10pt}
\caption{
\small
Local effort evidence for P0161 (MCI) and P0087 (MCI). Observed speech duration (P0161) and character count (P0087) are shown alongside their movement-conditioned residuals; darker highlighting denotes larger values within each panel.
}
\vspace{-10pt}
\label{fig:sample-effort}
\end{figure}

\noindent\textbf{Effort evidence.}
Figure~\ref{fig:sample-effort} shows two examples of localized effort evidence.
For participant P0161, several segments retain elevated effort residuals compared with other segments containing similar numbers of movements, despite a largely typical route.
For participant P0087, the trajectory contains several irregular route choices, yet elevated effort residuals remain confined to only a small subset of segments.
This contrast illustrates that action-side deviations and effort-side evidence need not co-occur.

\noindent\textbf{Dataset-level action localization.}
The examples above illustrate how local action evidence can highlight specific task decisions in individual traces.
We next test whether these patterns recur systematically across participants.
For participant \(j\) and action \(i\), we define action unexpectedness as \(U^A_{ji}=-\log \pi_{\hat{\eta}_j}(a_{ji}\mid s_{ji})\).

We evaluate it against two task-defined anchors derived independently from the grounded trace and store layout.
A \emph{route-divergence onset} is a movement that increases the shortest-path distance to the next observed task completion, while a \emph{target-opportunity bypass} occurs when a participant moves away from an unfinished target shelf despite an available purchase action.
We compute AUC within each participant and macro-average across participants, with participant-bootstrap confidence intervals.

\begin{table}[h]
\centering
\setlength{\tabcolsep}{5pt}
\renewcommand{\arraystretch}{1.0}
\caption{
Dataset-level action localization.
Top-\(10\%\) denotes event rate among each participant's most unexpected actions; enrichment is relative to baseline.
}
\label{tab:action-localization}
\scalebox{0.95}{
\begin{tabular}{@{}lcccc@{}}
\toprule
\textbf{Behavioral anchor}
& \textbf{Macro AUC}
& \textbf{Top-\(10\%\)}
& \textbf{Baseline}
& \textbf{Enrichment} \\
\midrule
Route-divergence onset
& 0.902
& 29.2\%
& 4.9\%
& 5.95$\times$ \\

Target-opportunity bypass
& 0.983
& 96.2\%
& 27.8\%
& 3.47$\times$ \\
\bottomrule
\end{tabular}}
\end{table}

For route-divergence onsets, action unexpectedness achieves a macro AUC of \(0.902\) and a \(5.95\times\) top-decile enrichment.
Target-opportunity bypasses show a consistent pattern, with a macro AUC of \(0.983\) and \(3.47\times\) enrichment.
Thus, the localization patterns illustrated above recur systematically across the dataset.

\subsection{Cognitive Status Classification}
\label{sec:cognitive-classification}
\vspace{-5pt}

Beyond factor recovery and localization, we test whether FIDM features carry predictive information about cognitive status, both as a standalone representation and when combined with each baseline.
Table~\ref{tab:classification} summarizes binary and three-class results.

\begin{table*}[t]
\centering
\setlength{\tabcolsep}{3pt}
\renewcommand{\arraystretch}{1}
\caption{
\small
Cognitive-status classification performance (\%) for binary and three-class settings.
For three-class classification, AUC is macro-averaged across normal cognition,
MCI, and major NCD.
Gray shading marks the proposed FIDM representation, teal indicates improvement
over the corresponding baseline, and bold underlined values indicate the best result
in each column.
}
\label{tab:classification}
\scalebox{0.95}{
\begin{tabular}{@{}lcccccccc@{}}
\toprule
& \multicolumn{4}{c}{\textbf{Binary}}
& \multicolumn{4}{c}{\textbf{Three-class}} \\
\cmidrule(lr){2-5}
\cmidrule(lr){6-9}

& \multicolumn{2}{c}{\textbf{5-fold CV}}
& \multicolumn{2}{c}{\textbf{Held-out Test}}
& \multicolumn{2}{c}{\textbf{5-fold CV}}
& \multicolumn{2}{c}{\textbf{Held-out Test}} \\
\cmidrule(lr){2-3}
\cmidrule(lr){4-5}
\cmidrule(lr){6-7}
\cmidrule(lr){8-9}

\textbf{Method}
& AUC & B-Acc
& AUC & B-Acc
& AUC & B-Acc
& AUC & B-Acc \\
\midrule

\multicolumn{9}{@{}l}{\textit{Clinical Baselines}} \\
\addlinespace[2pt]

GSDT score (1)
& 62.09
& 64.85
& 62.93
& 62.19
& 63.98
& \underline{\textbf{51.82}}
& 68.46
& 64.38 \\

GSDT score + subscores (8)
& 66.71
& 62.29
& 60.23
& 63.02
& 66.82
& 50.73
& 55.29
& 51.70 \\

\midrule
\multicolumn{9}{@{}l}{\textit{Behavioral and Language Baselines}} \\
\addlinespace[2pt]

Trajectory statistics (14)
& 62.40
& 61.00
& 67.02
& 63.40
& 56.41
& 41.09
& 64.17
& 44.36 \\

Frozen BERT (768)
& 67.64
& 62.50
& 70.33
& 61.07
& 70.38
& 49.50
& 70.80
& 44.40 \\

Frozen MacBERT (768)
& 65.85
& 61.38
& 67.35
& 66.56
& 69.48
& 49.97
& 70.64
& 46.72 \\

\midrule
\multicolumn{9}{@{}l}{\textit{Proposed Representation}} \\
\addlinespace[2pt]

\rowcolor{gray!15}
\textbf{FIDM features (5)}
& 70.39
& 65.12
& 66.79
& 64.74
& 67.39
& 49.65
& 67.85
& 56.34 \\

\midrule
\multicolumn{9}{@{}l}{\textit{Combinations with FIDM}} \\
\addlinespace[2pt]

GSDT score + FIDM (6)
& \textcolor{teal}{72.70}
& \textcolor{teal}{\underline{\textbf{67.29}}}
& \textcolor{teal}{71.35}
& \textcolor{teal}{68.56}
& \textcolor{teal}{70.16}
& 48.16
& \textcolor{teal}{\underline{\textbf{78.13}}}
& \textcolor{teal}{\underline{\textbf{71.72}}} \\

GSDT score + subscores + FIDM (13)
& \textcolor{teal}{\underline{\textbf{74.71}}}
& \textcolor{teal}{66.94}
& \textcolor{teal}{70.33}
& \textcolor{teal}{69.40}
& \textcolor{teal}{69.04}
& \textcolor{teal}{50.87}
& \textcolor{teal}{65.58}
& \textcolor{teal}{59.86} \\

Trajectory statistics + FIDM (19)
& \textcolor{teal}{69.46}
& \textcolor{teal}{65.70}
& \textcolor{teal}{68.74}
& \textcolor{teal}{65.91}
& \textcolor{teal}{64.40}
& \textcolor{teal}{46.31}
& 63.94
& \textcolor{teal}{56.72} \\

BERT + FIDM (773)
& \textcolor{teal}{72.46}
& \textcolor{teal}{63.28}
& \textcolor{teal}{\underline{\textbf{75.63}}}
& \textcolor{teal}{67.91}
& \textcolor{teal}{72.27}
& \textcolor{teal}{49.57}
& 68.91
& 40.41 \\

MacBERT + FIDM (773)
& \textcolor{teal}{72.61}
& \textcolor{teal}{65.74}
& \textcolor{teal}{74.88}
& \textcolor{teal}{\underline{\textbf{72.23}}}
& \textcolor{teal}{\underline{\textbf{72.61}}}
& 49.09
& \textcolor{teal}{71.12}
& \textcolor{teal}{54.32} \\

\bottomrule
\end{tabular}}
\vspace{-6pt}
\end{table*}

\noindent\textbf{Binary classification.}
FIDM uses only five parameters per participant and achieves the strongest standalone cross-validation performance, with an AUC of \(70.39\%\) and balanced accuracy of \(65.12\%\).
Adding FIDM improves every baseline across all four metrics.
The best combination varies by metric: BERT+FIDM achieves the highest held-out AUC (\(75.63\%\)), MacBERT+FIDM the highest held-out balanced accuracy (\(72.23\%\)), and GSDT score with subscores+FIDM the highest cross-validation AUC (\(74.71\%\)).
These gains show that the FIDM features add predictive information beyond each baseline representation.

\noindent\textbf{Three-class classification.}
FIDM alone achieves a held-out macro AUC of \(67.85\%\) and balanced accuracy of \(56.34\%\).
The strongest held-out results are obtained by GSDT score+FIDM, reaching \(78.13\%\) macro AUC and \(71.72\%\) balanced accuracy.
Fusion gains are less uniform than in the binary setting: FIDM consistently improves the held-out GSDT-score representations, while gains for trajectory and language representations depend on the metric.



\noindent\textbf{Factor contributions.} 
We isolate the contributions of the action and effort factor by evaluating each alone and in combination under both classification settings (Table~\ref{tab:ablation}). 
Under binary labels, the action factor alone outperforms the effort factor alone, while their combination yields the highest AUC and balanced accuracy.
Under three-class labels, the pattern reverses: the effort factors alone outperform the action factor alone.
The combined factors achieve the highest macro AUC, while the effort factors alone achieve the highest balanced accuracy.
These results support that action and effort capture complementary aspects of task execution. 

\begin{table}[h]
\centering
\setlength{\tabcolsep}{8pt}
\renewcommand{\arraystretch}{1}
\caption{
\small
Factor ablation under five-fold cross-validation (\%).
Three-class AUC is macro-averaged.
Gray shading marks the complete FIDM representation, and bold indicates the
best result in each column.
}
\label{tab:ablation}
\scalebox{0.9}{
\begin{tabular}{@{}lcccc@{}}
\toprule
& \multicolumn{2}{c}{\textbf{Binary}}
& \multicolumn{2}{c}{\textbf{Three-Class}} \\
\cmidrule(lr){2-3}
\cmidrule(lr){4-5}
\textbf{Representation}
& AUC
& B-Acc
& Macro AUC
& B-Acc \\
\midrule

Action feature (1)
& 67.86
& 63.67
& 62.62
& 47.86 \\

Effort features (4)
& 64.18
& 61.82
& 62.97
& \textbf{51.66} \\

\rowcolor{gray!15}
\textbf{Action + effort features (5)}
& \textbf{70.39}
& \textbf{65.12}
& \textbf{67.39}
& 49.65 \\

\bottomrule
\end{tabular}}
\end{table}
\vspace{-5pt}
\section{Discussion and Limitations}
\label{sec:discussion}
\vspace{-5pt}

The validation experiments support treating action choice and execution effort as distinct aspects of task performance. In the matched-observable setting, task executions with similar aggregate summaries can still differ in these two aspects, and FIDM retains this distinction.


The localization analyses connect participant-level parameters to specific portions of task execution. 
Model-relative unexpectedness is defined with respect to the fitted task model and the available alternatives, and therefore need not coincide with visually apparent deviations in individual traces. At the dataset level, elevated action unexpectedness is systematically enriched at independently defined task deviations across participants. Cognitive-status classification provides a complementary view: the inferred factors add information beyond clinical, behavioral, and language representations in the binary setting, while the three-class results vary more across representations.

We identify three main limitations. First, the empirical evaluation is centered on a single cognitive-assessment task, and generalization to other forms of sequential behavior remains to be evaluated. Second, the study includes 400 participants, a cohort size comparable to established clinically collected datasets in this area~\cite{luz2021alzheimer, lanzi2023dementiabank}.
Larger cohorts would support more reliable analysis of population heterogeneity and finer-grained cognitive-status differences. Third, task-trajectory reconstruction relies on automatic grounding. A manual audit found 19 of 20 sampled trajectories to be semantically consistent, but grounding errors may still propagate to downstream inference.

\vspace{-5pt}
\section{Conclusion}
\label{sec:conclusion}
\vspace{-5pt}

We introduced FIDM, a factorized inverse decision model that separates how individuals select actions from how those actions are executed under a shared sequential-task model.
Across controlled validation and real-world cognitive-screening data, 
the resulting factors provide an interpretable characterization of action selection and execution effort, while carrying information complementary to conventional behavioral, clinical, and language representations.
The results support incorporating structured execution signals alongside action trajectories in inverse models of sequential behavior.


\bibliographystyle{unsrtnat}
\bibliography{references}

\appendix
\newpage
\section{Task and Model Details}
\label{app:task-details}

\subsection{HK-GSDT Task Model}

The HK-GSDT requires participants to verbally direct an assessor-controlled figure through a fixed supermarket layout while completing a sequence of shopping goals.
The modeled task proceeds from the store entrance through item collection, an assessor-introduced bakery request, checkout, and exit.

FIDM represents the task as a deterministic grid world shared across participants.

\begin{table}[h]
\centering
\setlength{\tabcolsep}{6pt}
\renewcommand{\arraystretch}{1.05}
\caption{Components of the HK-GSDT task model used by FIDM.}
\label{tab:task-model}
\begin{tabular}{@{}lp{0.72\linewidth}@{}}
\toprule
\textbf{Component} & \textbf{Definition} \\
\midrule
Location
& Current walkable grid cell. \\

Task progress
& Completion status of task events needed to determine the remaining goals. \\

Phase
& Initial shopping, bakery retrieval, or checkout. \\

Actions
& Legal grid movements and task-completion actions available in the current state. \\

Transition
& Deterministic update of location, task progress, and phase following an action. \\

Task reference
& Phase-specific objective shared across participants; soft value iteration produces the action values used by the action factor. \\
\bottomrule
\end{tabular}
\end{table}

The phase variable is task-defined and specifies the active objective; it is not inferred from participant behavior.
Consequently, the same movement can receive different action values under different active goals.

\noindent\textbf{Reward and value specification.}
The task reward is defined with respect to the active phase, as
summarized in Table~\ref{tab:task_reward}. Target purchases use a unit
positive reward, while small movement and purchase penalties discourage
unnecessary actions. During checkout, the PAY reward depends on the
fraction of task targets completed. We use a discount factor
$\gamma=0.98$ and Bellman temperature $\tau_Q=0.05$ throughout.

\begin{table}[h]
\centering
\caption{Reward specification for the HK-GSDT task model.}
\label{tab:task_reward}
\begin{tabular}{lc}
\toprule
Action or event & Reward \\
\midrule
Movement (N/S/E/W) & $-0.02$ \\
First purchase of active navigation target & $+1.0$ \\
First bakery purchase in bakery phase & $+1.0$ \\
Repeated purchase of the same target & $-0.1$ \\
Purchase of a non-target item & $-0.2$ \\
First purchase of an out-of-phase target & $0$ \\
PAY in checkout phase & $0.5+1.5\,C(s)$ \\
PAY before checkout phase & $0$ \\
\bottomrule
\end{tabular}
\end{table}

Here,
\begin{equation}
C(s)
=
\frac{
n_{\mathrm{nav}}+
\mathbb{I}(\mathrm{bakery\ purchased})
}{5},
\end{equation}
where $n_{\mathrm{nav}}$ is the number of completed navigation targets
among vegetables, meat, egg/tofu, and seasoning. The OUT action is not
included in the action likelihood and therefore receives no reward.

\subsection{Transcript Grounding and Effort Alignment}

Each timestamped participant--assessor dialog is converted into an ordered task-execution trace.
For each route-description segment, the grounding model receives the current grid location, preceding dialog context, and verbal route description and returns an ordered sequence of grid movements.
Deterministic replay reconstructs the visited states, while grounded purchase and payment actions update task progress.

Grounding was performed using a locally deployed DeepSeek-v4 Flash model.
Due to IRB constraints on protected health information, all transcript processing remained on local infrastructure, restricting grounding to locally deployable models.

Grid transitions and speech segments are not in one-to-one correspondence: a route-description segment may encode several consecutive movements.
Grounded movements and task-completion actions, including BUY and PAY, enter the action likelihood. Route-description segments may contain several consecutive movements, while execution observations are measured once per segment and conditioned on the number of aligned movements.

\paragraph{Grounding quality audit.}
All 400 task-execution traces used in the study passed automated structural-consistency checks.
In a manual audit of 20 transcript--trajectory pairs, 19 were semantically consistent, while one contained a route completion not sufficiently supported by the transcript.

\subsection{Effort Likelihoods}

Let segment $u$ of participant $j$ contain $n_{ju}$ grounded movements and
\[
x_{ju}=\log(1+n_{ju}).
\]
For execution channel $k$,
\[
\xi_{juk}
=
\alpha_k+\beta_k x_{ju}+\theta_{jk},
\]
where $\alpha_k$ and $\beta_k$ are population-level parameters estimated from the training partition and $\theta_{jk}$ is participant specific.

\begin{table}[h]
\centering
\setlength{\tabcolsep}{5pt}
\renewcommand{\arraystretch}{1.08}
\caption{Parameterization of the four execution-channel likelihoods.}
\label{tab:effort-likelihoods}
\begin{tabular}{@{}llll@{}}
\toprule
\textbf{Channel}
& \textbf{Observation}
& \textbf{Likelihood}
& \textbf{Conditional parameter} \\
\midrule
Speech duration
& $d_{ju}>0$
& $\mathrm{Gamma}(\mu^{(d)}_{ju},\kappa_d)$
& $\log\mu^{(d)}_{ju}=\xi_{ju,d}$ \\

Character count
& $c_{ju}\geq0$
& $\mathrm{NB}(\mu^{(c)}_{ju},\kappa_c)$
& $\log\mu^{(c)}_{ju}=\xi_{ju,c}$ \\

Dialog turns
& $r_{ju}\geq0$
& $\mathrm{NB}(\mu^{(r)}_{ju},\kappa_r)$
& $\log\mu^{(r)}_{ju}=\xi_{ju,r}$ \\

Positive pause
& $p_{ju}\geq0$
& $\mathrm{Bernoulli}+\mathrm{LogNormal}$
& $\mathbb{E}[\log p_{ju}\mid p_{ju}>0]=\xi_{ju,p}$ \\
\bottomrule
\end{tabular}
\end{table}

The Gamma likelihood is parameterized by mean $\mu$ and shared shape $\kappa$.
For the negative-binomial channels,
\[
\operatorname{Var}(e)
=
\mu+\frac{\mu^2}{\kappa}.
\]
Pause presence and positive pause duration are modeled separately.
The participant-specific pause parameter is estimated from the positive-duration component.

For localization, residuals are computed relative to the population-level movement-conditioned expectation.
For speech duration,
\[
R^{\mathrm{time}}_{ju}
=
\log d_{ju}
-
\left[
\alpha_d+\beta_d\log(1+n_{ju})
\right],
\]
and for the count channels,
\[
R^{(k)}_{ju}
=
\log(1+e^{(k)}_{ju})
-
\log(1+\widehat{\mu}^{(k)}_{ju}),
\qquad
k\in\{\mathrm{char},\mathrm{turn}\},
\]
where
$\widehat{\mu}^{(k)}_{ju}
=
\exp[\alpha_k+\beta_k\log(1+n_{ju})]$
is the population-level movement-conditioned expectation.
These residuals are used for localization and visualization; participant-level effort parameters are estimated from the likelihoods above.

\section{Experimental Details}
\label{app:experimental-details}

\subsection{Representations}

The trajectory baseline contains 14 participant-level summary statistics:
total action count, movement count, BUY count, PAY and OUT indicators, counts of northward, southward, eastward, and westward movements, direction changes, immediate backtracks, unique cells visited, revisited steps, and a state--movement consistency indicator.

The language baselines use frozen
\texttt{google-bert/bert-base-chinese} and
\texttt{hfl/chinese-macbert-base} encoders with participant-only utterances.
Token representations are mean-pooled within each input chunk and combined across chunks by a token-length-weighted average, yielding one 768-dimensional representation per participant.
The model checkpoints are available at
\url{https://huggingface.co/google-bert/bert-base-chinese}
and
\url{https://huggingface.co/hfl/chinese-macbert-base}.

\subsection{Feature Fusion and Classification}

For fusion with language embeddings, missing values in the low-dimensional feature block are median-imputed and each feature block is standardized separately.
Each standardized block is then scaled by the inverse square root of its dimensionality,
\[
\widetilde{x}^{(b)}
=
\frac{\operatorname{standardize}(x^{(b)})}{\sqrt{d_b}},
\]
after which the blocks are concatenated.
All preprocessing and parameters are fitted on the corresponding training data and, for cross-validation, separately in each training fold.

All downstream experiments use the same L2-regularized logistic
regression classifier with $C=1$ and
\texttt{class\_weight=balanced}, without representation-specific
classifier hyperparameter tuning. For held-out evaluation, the full pipeline is
refitted on all 332 training participants and evaluated on the
68-participant test set.

\subsection{Task-Model Hyperparameter Selection}

The core reward values for target purchases, repeated and irrelevant
purchases, and checkout were specified manually and checked for
non-degenerate task policies. We selected the movement cost and
Bellman temperature from
\[
c_{\mathrm{move}}\in\{-0.01,-0.02,-0.05\},
\qquad
\tau_Q\in\{0.03,0.05,0.08,0.12\},
\]
using five-fold cross-validation on the training partition. The
selected setting, $c_{\mathrm{move}}=-0.02$ and $\tau_Q=0.05$, was
then fixed for held-out evaluation.
\section{Validation Details}
\label{app:validation-details}

\subsection{Controlled Factor Recovery}

Semi-synthetic traces use the observed HK-GSDT task structure.
Action sensitivity, speech duration, character count, dialog-turn count, and positive pause duration are manipulated independently.
Action sequences are sampled from the task-value policy, and execution observations are generated from the corresponding likelihood families.
For each recovered parameter, we report the standardized partial slope from a regression on all five generative manipulations.

The missing-observation condition removes 30\% of effort observations before inference.
A separate increased-noise condition evaluates recovery under greater observation variability.

\subsection{Matched Observable Behavior}

Each A/B pair uses the same observed HK-GSDT state--value contexts and identical total movement exposure.
Condition A uses lower action sensitivity and the original effort segmentation, whereas condition B uses higher action sensitivity and pairwise-merged segments.
Execution-channel offsets are chosen so that the corresponding raw aggregate totals match across conditions in expectation.
Both members of each pair are assigned to the same cross-validation fold.

Across five repetitions, the maximum absolute standardized mean difference among the matched raw and per-movement summaries is $0.072$.

\begin{table}[h]
\centering
\setlength{\tabcolsep}{6pt}
\renewcommand{\arraystretch}{1.04}
\caption{
Factor recovery in the matched-observable experiment.
Mean differences are condition B minus condition A.
}
\label{tab:matched-recovery}
\begin{tabular}{@{}lccc@{}}
\toprule
\textbf{Recovered factor}
& \textbf{Pearson $r$}
& \textbf{Pairwise direction}
& \textbf{Mean B--A} \\
\midrule
Action $\hat{\eta}$      & 0.566 & 0.924 & 12.766 \\
Time $\hat{\theta}$      & 0.615 & 0.866 & 0.348 \\
Character $\hat{\theta}$ & 0.652 & 0.888 & 0.376 \\
Turns $\hat{\theta}$     & 0.737 & 0.936 & 0.590 \\
Pause $\hat{\theta}$     & 0.538 & 0.819 & 0.520 \\
\bottomrule
\end{tabular}
\end{table}


\subsection{Dataset-Level Action Localization}


A \emph{route-divergence onset} is a movement after which the shortest-path distance to the next observed task completion is greater than immediately before the movement.
A \emph{target-opportunity bypass} occurs when a participant moves away from an unfinished target shelf despite an available purchase action.

AUC is computed within participant and macro-averaged across participants containing both positive and negative instances.
Participant-bootstrap confidence intervals quantify uncertainty.
Top-decile enrichment compares the event rate among each participant's most unexpected 10\% of actions with the corresponding baseline event rate.

\begin{table}[h]
\centering
\setlength{\tabcolsep}{6pt}
\renewcommand{\arraystretch}{1.04}
\caption{Additional statistics for dataset-level action localization.}
\label{tab:action-localization-app}
\begin{tabular}{@{}lcc@{}}
\toprule
\textbf{Behavioral anchor}
& \textbf{Macro AUC (95\% CI)}
& \textbf{Top-decile enrichment} \\
\midrule
Route-divergence onset
& 0.902 [0.877, 0.925]
& 5.95$\times$ \\

Target-opportunity bypass
& 0.983 [0.962, 1.000]
& 3.47$\times$ \\
\bottomrule
\end{tabular}
\end{table}

The two anchors are task-defined behavioral references, with the next observed task completion used retrospectively to define the route reference.


\end{document}